\documentclass[10pt,twocolumn,letterpaper]{article}

\usepackage[pagenumbers]{wacv} 

\definecolor{wacvblue}{rgb}{0.21,0.49,0.74}
\usepackage[pagebackref,breaklinks,colorlinks,allcolors=wacvblue]{hyperref}
\usepackage{multirow}
\def\wacvPaperID{52} 
\def\confName{WACV}
\def\confYear{2026}

\title{Objectomaly: Objectness-Aware Refinement for OoD Segmentation with Structural Consistency and Boundary Precision}

\author{
Jeonghoon Song$^{1}$ \quad Sunghun Kim$^{2}$ \quad Jaegyun Im$^{2}$ \quad Byeongjoon Noh$^{2}$\\
$^{1}$Department of Future Convergence Technology, Soonchunhyang University\\
$^{2}$Department of AI and Big Data, Soonchunhyang University\\
22 Soonchunhyang-ro, Asan 31538, South Korea\\
{\tt\small \{sjhon121215, ksh0816, imij0522, powernoh\}@sch.ac.kr} \\
}

\begin{document}
\maketitle
\vspace*{-2.6em}
\begin{abstract}
Out-of-Distribution (OoD) segmentation is critical for safety-sensitive applications like autonomous driving. However, existing mask-based methods often suffer from boundary imprecision, inconsistent anomaly scores within objects, and false positives from background noise. We propose \textbf{\textit{Objectomaly}}, an objectness-aware refinement framework that incorporates object-level priors. Objectomaly consists of three stages: (1) Coarse Anomaly Scoring (CAS) using an existing OoD backbone, (2) Objectness-Aware Score Calibration (OASC) leveraging SAM-generated instance masks for object-level score normalization, and (3) Meticulous Boundary Precision (MBP) applying Laplacian filtering and Gaussian smoothing for contour refinement. Objectomaly achieves state-of-the-art performance on key OoD segmentation benchmarks, including SMIYC AnomalyTrack/ObstacleTrack and RoadAnomaly, improving both pixel-level (AuPRC: 96.99, FPR$_{95}$: 0.07) and component-level (F1-score: 83.44) metrics. Ablation studies and qualitative results on real-world driving videos further validate the robustness and generalizability of our method. Implementation details are available at: \url{https://github.com/hon121215/Objectomaly}
\end{abstract}

\section{Introduction}Semantic segmentation is a fundamental computer vision task that assigns a semantic class label to each pixel, playing a key role in applications such as autonomous driving~\cite{noh2020vision, noh2022novel, wang2025semflow}, aerial imagery~\cite{ji2024unleashing}, and surveillance~\cite{wang2025toward, lee2025saliencymix+}. While deep learning-based models have achieved high accuracy on In-Distribution (ID) objects, they often fail to generalize to Out-of-Distribution (OoD) objects—semantic classes unseen during training. In safety-critical domains like autonomous driving, such failures can result in severe consequences, e.g., missing unexpected obstacles such as animals or debris~\cite{vobecky2022drive, bogdoll2022anomaly}.

To clarify our scope, we distinguish OoD segmentation from Anomaly Segmentation (AS) and Uncertainty Quantification (UQ). OoD segmentation aims to localize novel semantic classes, whereas AS focuses on rare or irregular patterns, and UQ quantifies model uncertainty~\cite{zhao2024segment, grcic2022densehybrid, schweighofer2023quantification}. Our work specifically addresses OoD segmentation.

Recent OoD approaches primarily rely on pixel-wise anomaly scores~\cite{vojivr2024pixood, sodano2024open}, but often suffer from three structural issues: imprecise boundaries, false positives due to textured backgrounds, and inconsistent anomaly scores within objects~\cite{wu2023conditional, Ma_2025_WACV, Ackermann_2023_BMVC}. While object-aware methods such as RbA~\cite{nayal2023rba}, Maskomaly~\cite{Ackermann_2023_BMVC}, and Mask2Anomaly~\cite{rai2023unmasking} improve semantic awareness, most still rely heavily on softmax confidence scores, which are prone to overconfidence and spatial inconsistency.

In this paper, we propose \textbf{\textit{Objectomaly}}, a post-hoc OoD segmentation method that integrates object-level cues into a structured refinement process. Objectomaly consists of three stages: 1) Coarse Anomaly Scoring (CAS) using an existing model like Mask2Anomaly; 2) Objectness-Aware Score Calibration (OASC), which aligns scores using class-agnostic masks from the Segment Anything Model (SAM); and 3) Meticulous Boundary Precision (MBP), which refines boundaries using Laplacian filtering and Gaussian smoothing. This design ensures both semantic precision and structural consistency without additional training.

We evaluate Objectomaly on four OoD benchmarks—SMIYC AnomalyTrack (AT)/ObstacleTrack (OT)~\cite{chan2021segmentmeifyoucan}, and RoadAnomaly (RA)~\cite{lis2019detecting}. Our method achieves state-of-the-art (SOTA) performance across pixel-level and component-level metrics, and qualitative results confirm its effectiveness in real-world driving scenarios.

\section{Related Work}\subsection{Image Segmentation}
Image segmentation tasks—semantic, instance, and panoptic—have evolved from pixel-wise classification to unified object-level formulations. Early CNN-based models such as DeepLab~\cite{chen2018encoder}, PSPNet~\cite{zhao2017pyramid}, and U-Net~\cite{ronneberger2015u} enabled pixel-accurate predictions but suffered from limited contextual understanding due to the locality of convolutional operations~\cite{zheng2021rethinking}.

Transformer-based models addressed this issue by modeling long-range dependencies. SETR~\cite{zheng2021rethinking} and Swin Transformer~\cite{liu2021swin} introduced global attention and hierarchical structure, while frameworks such as MaskFormer~\cite{cheng2021per} and Mask2Former~\cite{cheng2022masked} unified semantic and instance segmentation via set prediction and object-centric embeddings. This shift laid the foundation for panoptic segmentation~\cite{xiao20243d,Chen_2023_ICCV} and general-purpose segmenters like SAM~\cite{kirillov2023segment}, which enable open-vocabulary and prompt-driven segmentation.

However, most models assume a closed-set label space, limiting their applicability in open-world scenarios where unexpected objects frequently appear. This motivates the OoD segmentation task, which aims to detect and localize novel entities beyond the training distribution~\cite{gao2024generalize,li2023open}.

\subsection{OoD Segmentation Problem}

Initial approaches to OoD segmentation leveraged pixel-level uncertainty scores derived from softmax outputs~\cite{fort2021exploring,liang2018enhancing}, entropy, or calibration. Despite their simplicity, these methods produce fragmented masks and exhibit poor boundary precision due to their lack of spatial structure modeling.

Recent advances pivot to mask-based frameworks that assign anomaly scores at the object level, improving spatial coherence and semantic reliability~\cite{rai2023unmasking,liu2023residual}. Mask2Anomaly~\cite{rai2023unmasking} extends Mask2Former with contrastive learning-based anomaly scoring, while Maskomaly~\cite{Ackermann_2023_BMVC} performs zero-shot detection using intermediate feature rejection/acceptance without additional training.

To further improve boundary accuracy and robustness, methods like RbA~\cite{nayal2023rba} and UNO~\cite{Delic_2024_BMVC} introduce class-rejection strategies and uncertainty-aware scoring. However, many frameworks still rely on softmax-based classifiers and lack explicit objectness modeling or contour-level refinement~\cite{zhang2024csl,wang2023segrefiner}.

Our study builds on this trend by proposing a refinement strategy for mask-based OoD segmentation that enhances structural consistency through explicit alignment and boundary-aware post-processing, tailored for visually complex, real-world environments.
\section{Proposed Method}This section details the proposed Objectomaly method. We first provide a concise overview of Mask2Anomaly, which serves as our baseline model. We then identify its key structural limitations, which motivate the design of our object-aware refinement strategy.

\subsection{Preliminary}
\subsubsection{Background of Mask2Anomaly Method}
This section briefly describes Mask2Anomaly~\cite{rai2023unmasking}, which forms the foundation of our proposed method. Mask2Anomaly extends the conventional pixel-level anomaly segmentation approach to a mask-level framework based on the mask-transformer architecture, specifically Mask2Former~\cite{cheng2022masked}. The architecture comprises three main components: an encoder (backbone), a pixel decoder, and a transformer decoder. First, the encoder converts an input image \(I \in \mathbb{R}^{H \times W \times 3}\) into an abstract feature map. Then, the pixel decoder upsamples this map into high-resolution pixel embeddings. Finally, the transformer decoder employs fixed object queries (mask embeddings) to predict class scores \(C \in \mathbb{R}^{K}\) and corresponding class masks \(M \in [0,1]^{H \times W}\).

The mask-level anomaly score is computed as~\cite{rai2023unmasking}:
\begin{equation}
    S(x) = \text{softmax}(C)^\top \cdot \text{sigmoid}(M)
\end{equation}
\begin{equation}
    f(x) = 1 - \max_{z\in\mathcal{Z}} S_z(x),\quad \mathcal{Z}=\{1, \dots, Z\}
\end{equation}
where $\mathcal{Z}$ is the set of known semantic classes, and \(\max_{z\in\mathcal{Z}} S_z(x)\) is the highest confidence for the known class at each pixel. This allows \(f(x)\in[0,1]^{H\times W}\) to have a large value at points where no known class can adequately describe the pixel.
Unlike pixel-wise softmax approaches, this formulation computes anomaly scores at the mask level, providing a comprehensive assessment.

In addition, Mask2Anomaly integrates a Global Masked Attention (GMA) mechanism, explicitly differentiating foreground and background attention. GMA is defined as:
The mask-level anomaly score is computed as~\cite{rai2023unmasking}:
\begin{equation}
\begin{split}
  X_{\text{out}} ={}& \;\text{softmax}(M^F_l + QK^\top)V\\
                   &+\,\text{softmax}(M^B_l + QK^\top)V + X_{\text{in}}
\end{split}
\end{equation}
where \(Q, K, V \in \mathbb{R}^{N \times d}\) represent the query, key, and value matrices derived from feature embeddings, and \(X_{\text{in}}\) denotes the input features from the previous layer. As defined in~\cite{rai2023unmasking}, attention biases are conditioned on the previous mask \(M_{l-1}\): foreground bias \(M^F_l(i,j)\) assigns 0 if \(M_{l-1}(i,j) \geq 0.5\) and \(-\infty\) otherwise, while background bias \(M^B_l(i,j)\) assigns 0 if \(M_{l-1}(i,j) < 0.5\) and \(-\infty\) otherwise.

This approach explicitly separates foreground and background features, improving anomaly detection effectiveness. Mask2Anomaly also employs mask contrastive learning to enhance semantic distinction between normal and anomalous regions. (We refer to the preceding process as the semantic confidence estimation stage in Mask2Anomaly throughout the remainder of this paper.)

Finally, to refine initial anomaly scores, Mask2Anomaly conducts the refinement of the mask \(R_M\)~\cite{rai2023unmasking}:
\begin{equation}
    R_M = \bar{C} \cdot \bar{M}, \quad f_r(x) = R_M \odot f(x)
\end{equation}
where \(\bar{M}\) is a binary mask derived from thresholding sigmoid outputs at 0.5, and \(\bar{C}\) represents semantic class confidence~\cite{rai2023unmasking}:
\begin{equation}
\bar{C} =
\begin{cases}
1, & \arg\max(\text{softmax}(C)) \in \{\text{Things}, \text{Road}\} \\[1pt] & \text{ and } \max(\text{softmax}(C)) > 0.95 \\[3pt]
0, & \text{otherwise}
\end{cases}
\end{equation}

This refinement suppresses irrelevant background masks and produces the final anomaly score.

\subsubsection{Problem Statement}
Despite effectively addressing pixel-level false positives, the refinement mechanism of Mask2Anomaly, based on semantic confidence \(\bar{C}\) and binary masks \(\bar{M}\), introduces the following three critical structural limitations (see Fig~\ref{fig:limitation} for a visual description):

\begin{figure}[!t]
  \centering
  \subfloat[]{%
    \includegraphics[width=0.15\textwidth]{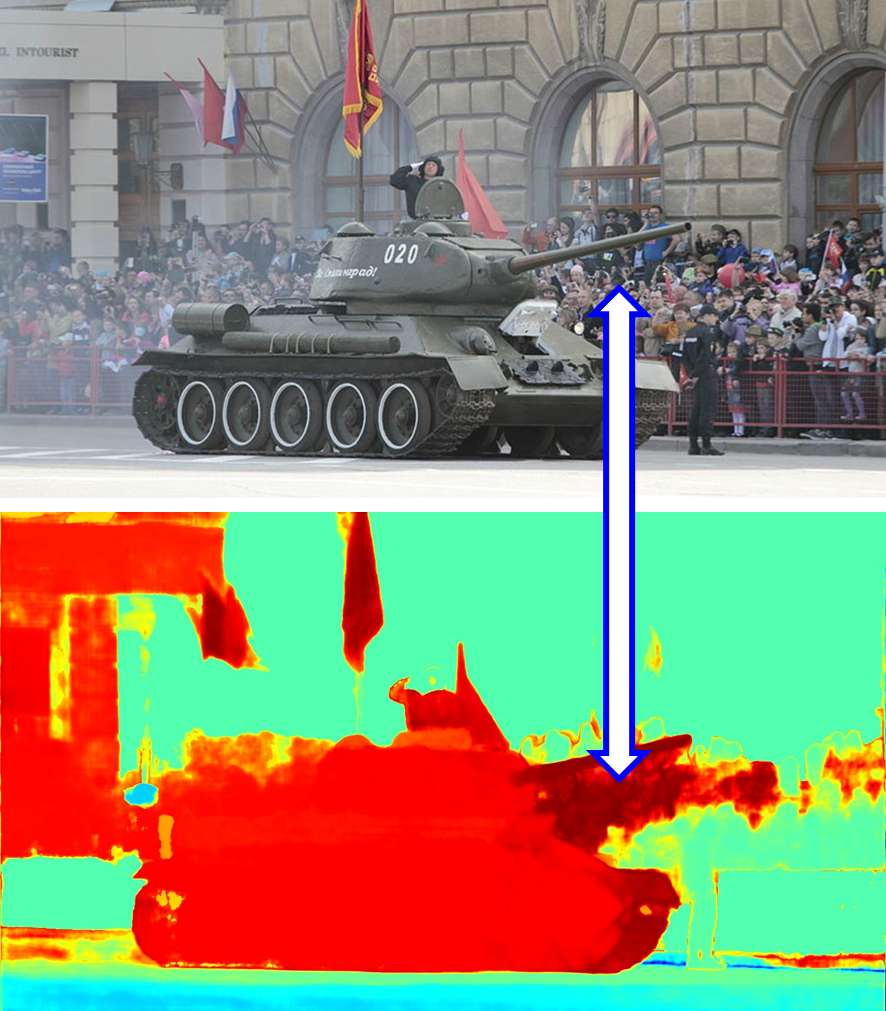}
    \label{subfig:limitation1-inaccurate-boundaries}
  }
  \subfloat[]{%
    \includegraphics[width=0.15\textwidth]{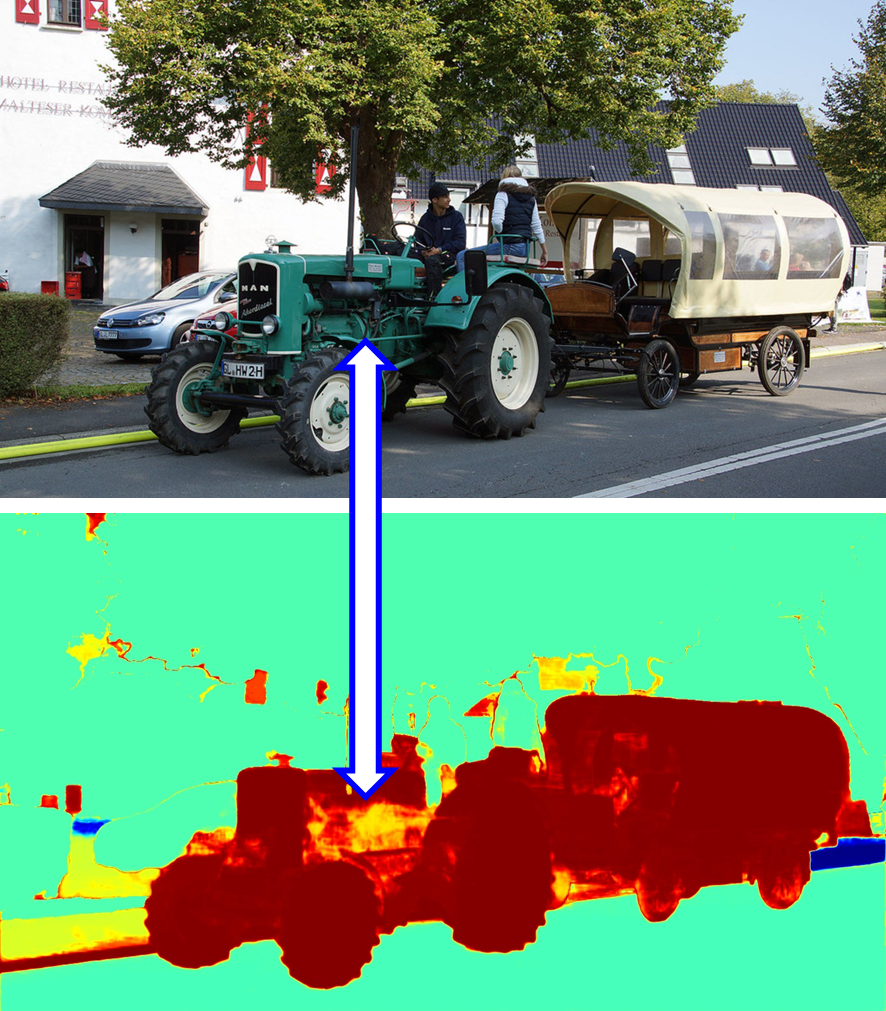}
    \label{subfig:limitation2-spatial-consistency}
  }
  \subfloat[]{%
    \includegraphics[width=0.15\textwidth]{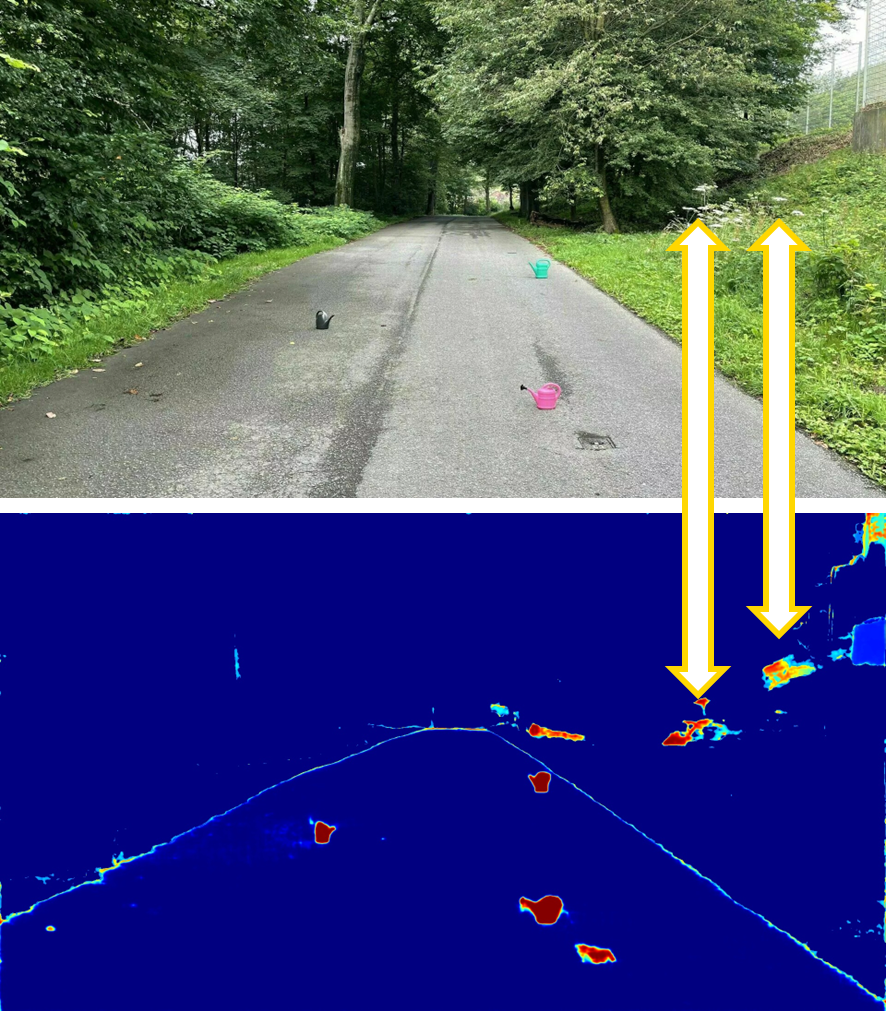}
    \label{subfig:limitation3-false-positives}
  }
  \caption{The limitations in current OoD segmentation method: (a) Inaccurate boundaries between adjacent objects, (b) Lack of spatial consistency within anomaly scores of the same object, (c) Increase in false positives due to background noise.}
  \label{fig:limitation}
  \vspace{-15pt}
\end{figure}
\begin{figure*}[!htbp]
  \centering
  \subfloat[]{%
    \includegraphics[width=0.95\textwidth]{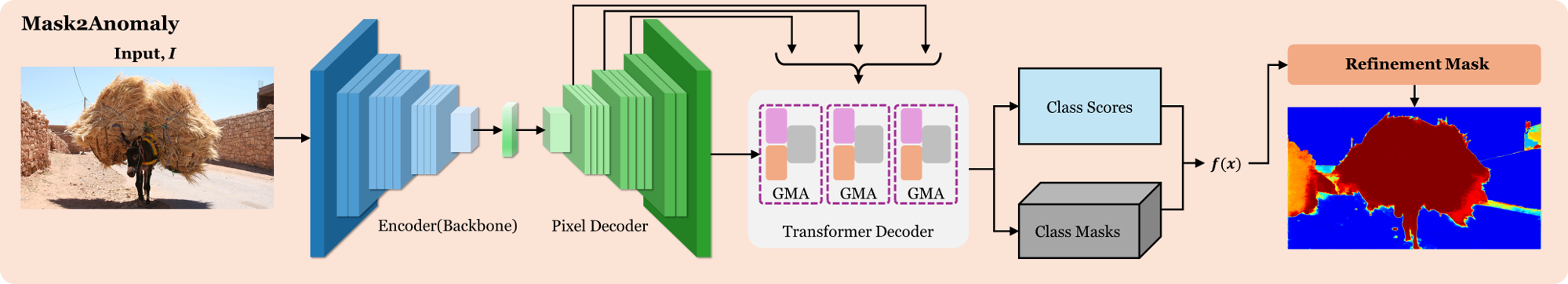}%
    \label{subfig:architecture-m2a}
  }\\
  \subfloat[]{%
    \includegraphics[width=0.95\textwidth]{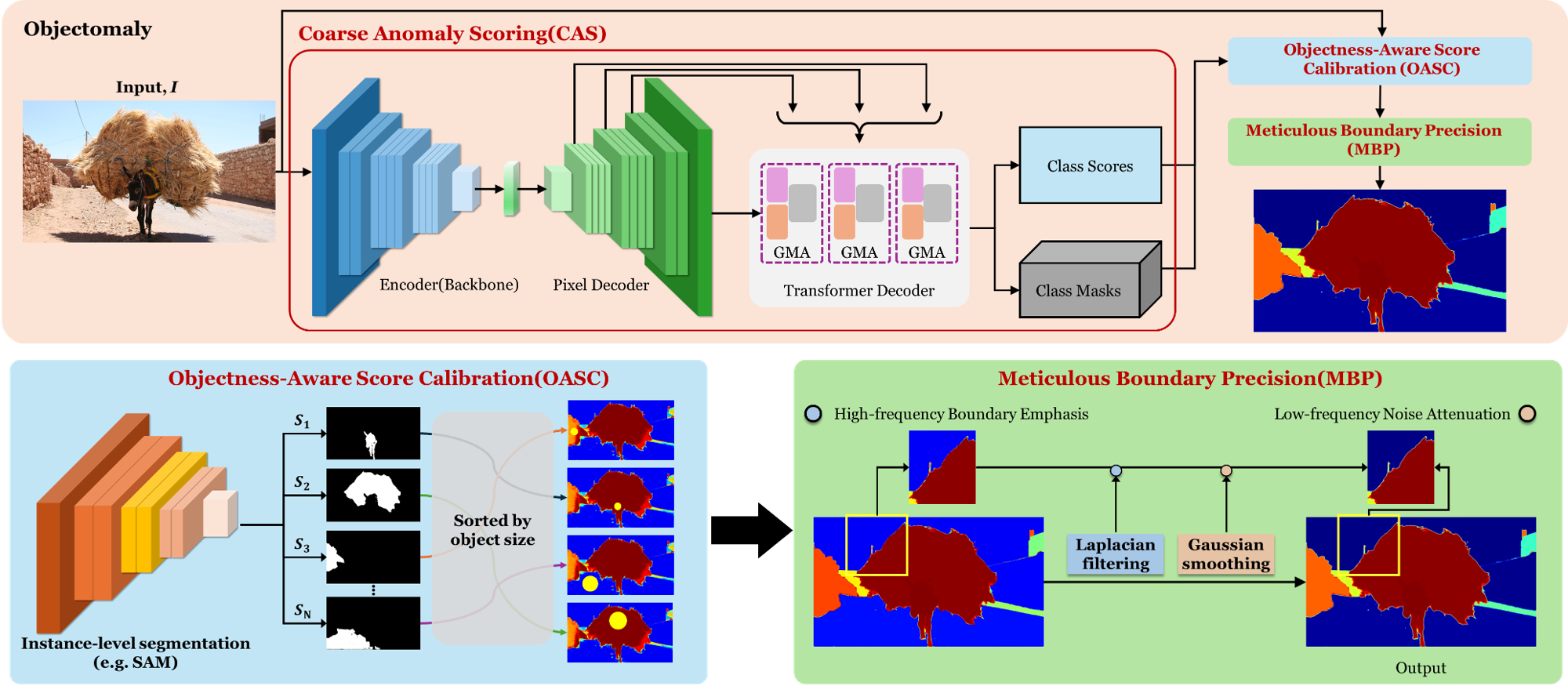}%
    \label{subfig:architecture}
  }
  \caption{Overview of Mask2Anomaly and the proposed Objectomaly. (a) Mask2Anomaly: An initial anomaly map is generated via semantic confidence estimation using mask-level softmax scores and global attention, followed by background suppression through a refinement mask. (b) Objectomaly: (1) CAS generates a coarse anomaly map using an OoD model; (2) OASC calibrates scores within SAM-derived instance masks to enhance spatial consistency and reduce noise; (3) MBP sharpens object boundaries via Laplacian filtering and Gaussian smoothing.}
  \label{fig:architecture}
  \vspace{-5pt}
\end{figure*}

\paragraph{\textbf{Inaccurate boundaries between adjacent objects.}} The reliance on pixel-level softmax scores leads to artificially inflated confidence values near foreground-background boundaries. This phenomenon results in blurred or ambiguous delineations, potentially causing adjacent objects to merge.

\paragraph{\textbf{Lack of spatial consistency within anomaly scores of the same object.}} Since the refinement mechanism depends on pixel-level semantic confidence and mask activation, inconsistency arises within anomaly scores of a single object. Even subtle variations in semantic confidence can fragment a continuous object into disconnected blobs. For instance, for two pixels \(x_i\) and \(x_j\) within the same object \(O_k\) (i.e.\ \(x_i, x_j \in O_k\)), differing confidence scores can lead to spatially disconnected anomaly representations.

\paragraph{\textbf{Increase in false positives due to background noise.}} Though the semantic confidence estimation model attempts to exclude low-confidence regions, background areas with complex textures, shadows, or other visual noise can produce transiently high semantic scores. This misclassification leads to these regions being mistaken for foreground objects, thereby increasing the rate of false positives.

These limitations are particularly detrimental in complex scenes, low-saliency objects, or ambiguous boundaries, leading to critical decision-making errors in autonomous systems. To address these issues, it is necessary to move beyond pixel-based and semantic-dependent refinement strategies towards an object-level structural refinement. Consequently, we propose the Objectness-Aware Mask Refinement approach, leveraging instance-level masks.

\subsection{The proposed Objectomaly method}
The Objectomaly method refines initial anomaly predictions generated by the semantic confidence estimation stage by incorporating instance-level structural information. The overall pipeline consists of three main stages: CAS, OASC, and MBP, as illustrated in Fig~\ref{fig:architecture}.

\subsubsection{Coarse Anomaly Scoring}
In the CAS stage, we generate an initial anomaly map, \(A_\text{init} \in [0, 1]^{H \times W}\) from the input image \(I\), using the semantic confidence estimation stage of the Mask2Anomaly model, denoted as \(P(\cdot)\), but bypassing its final refinement mask module:
\begin{equation}
    A_\text{init} = P(I)
\end{equation}
This initial map, reflecting the baseline model's raw semantic uncertainty, serves as the input for our subsequent structural refinement stages.

\subsubsection{Objectness-Aware Score Calibration}
In this stage, we structurally recalibrate the initial anomaly map (\(A_\text{init}\)) using object masks generated by the SAM. SAM roughly outputs a set of \(N\) binary object masks for a given image as follows:
\begin{equation}
    \mathcal{S} = \{S_1, S_2, \dots, S_N\}, \quad S_k \in \{0,1\}^{H \times W}
\end{equation}
However, because these masks can overlap, we process them in descending order of their size, \(|S_k|\), to create a prioritized set \({\mathcal{S}^{\prime}}\):

\begin{equation}
  |S_k|
  = \sum_{u=1}^{H}\sum_{v=1}^{W} S_k(u,v),
  \quad
  S' = \{S'_1,\,S'_2,\,\dots,\,S'_N\}.
\end{equation}
where, \(S_k(u,v)=1\) indicates that the pixel \((u,v)\) belongs to the object, whereas \(S_k(u,v)=0\) indicates that it belongs to the background.

We adopt SAM because of its distinct advantages in object-level segmentation. As a class-agnostic model, SAM generalizes strongly to unseen objects without predefined semantic labels~\cite{kirillov2023segment}, making it ideal for OoD segmentation. Moreover, its ability to generate high-quality instance masks with fine-grained boundaries provides a robust structural prior for calibrating the coarse anomaly scores. These capabilities make SAM a reliable source for extracting object-level regions that are not constrained by semantic categories, aligning well with the objective of anomaly detection in unknown regions.

\begin{figure}[!t]
    \centering
    \includegraphics[width=\linewidth]{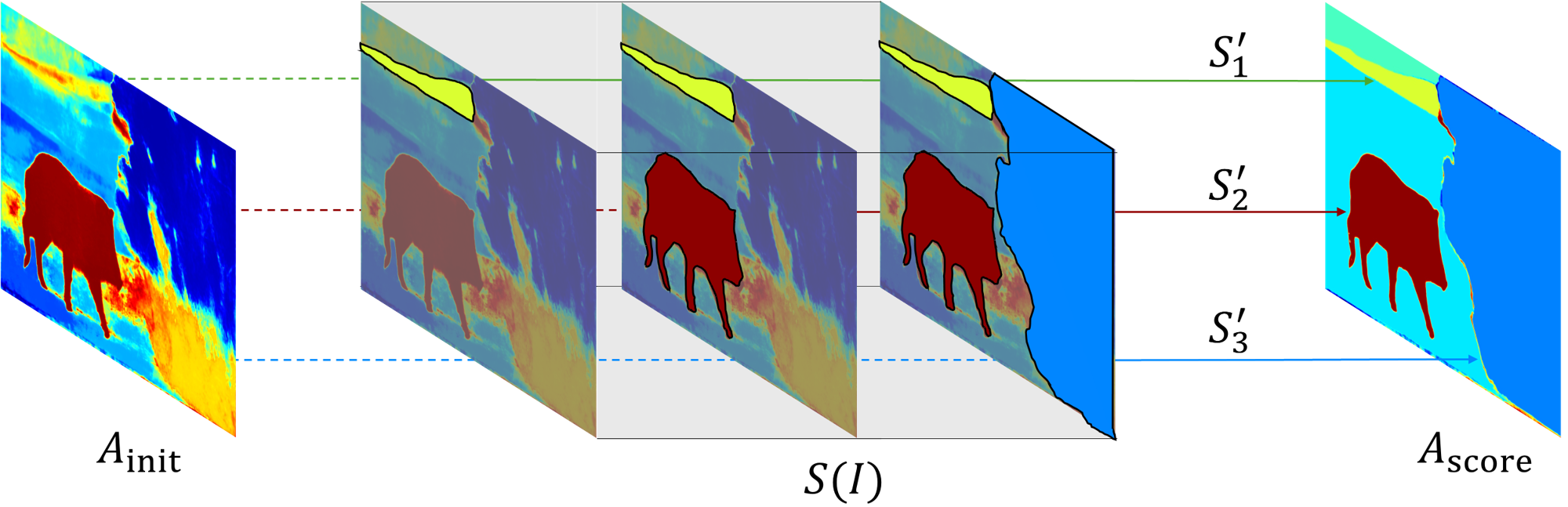}
    \caption{Illustration of the OASC process. Given the initial anomaly score map \( A_{\text{init}} \) from the Coarse Anomaly Scoring, class-agnostic segmentation masks \( S(I) \) generated by SAM serve as object masks. Refinement is performed progressively, starting with smaller objects and expanding to larger ones, ensuring that anomaly scores are adjusted within object boundaries. This process produces the refined anomaly score map \( A_{\text{score}} \), improving object-background separation and reducing false positives.}
    \label{fig:oasc}
    \vspace{-5pt}
\end{figure}

Fig~\ref{fig:oasc} visually illustrates how the initial anomaly score map \(A_\text{init}\) is integrated with object masks to perform objectness-aware refinement. For each mask \({S}^{\prime}_k\), we compute the representative anomaly score (\(\alpha_k\)) by averaging anomaly scores from \(A_\text{init}\) over the mask region:
\begin{equation}
  \alpha_k
  = \frac{1}{\lvert {S}^{\prime}_k\rvert}
    \sum\limits_{x \in {S}^{\prime}_k}
    A_{\mathrm{init}}(x) \,
\end{equation}

Subsequently, anomaly scores within each mask region are replaced by this averaged score, preserving scores outside mask regions:
\begin{align}
    A_{\text{score}}(x) =
    \begin{cases}
    \alpha_k, & \text{if } x \in {S}^{\prime}_k \\[3pt]
    A_\text{init}(x), & \text{otherwise}
    \end{cases}
\end{align}

This calibration removes pixel-wise score variability, improves spatial consistency within objects, alleviates structural distortions, and mitigates false positives and overconfidence near object boundaries. OASC thus ensures structural alignment, serving as a foundational step for the subsequent boundary refinement.

\subsubsection{Meticulous Boundary Precision}
Although OASC significantly improves structural consistency, local noise and subtle gradient variations may persist around object boundaries, especially in complex or low-contrast scenarios, potentially causing misclassification errors. 

To resolve this, we employ a gradient-based boundary refinement technique. Initially, we apply a Laplacian operator to the calibrated score map \(A_{\text{score}}\) to accentuate boundaries via second-order derivatives:

\begin{equation}
    A_{\text{lap}}(x) = \left| \frac{\partial^2 A_{\text{score}}}{\partial x^2} + \frac{\partial^2 A_{\text{score}}}{\partial y^2} \right|
\end{equation}

This step highlights sharp gradient changes corresponding to object boundaries. To mitigate the Laplacian's sensitivity to high-frequency noise, we then apply Gaussian smoothing with a kernel \(G_\sigma\):

\begin{equation}
    \tilde{A}_{\text{score}} = G\sigma * A_{\text{lap}}
\end{equation}
where \(\sigma\) controls the trade-off between boundary sharpness and noise suppression. This operation structurally refines object contours by smoothing local variations.

Thus, MBP is not merely post-processing; it is a critical module that exploits gradient properties to refine boundary precision. Importantly, MBP maintains soft anomaly score maps without thresholding, offering flexible downstream applications and preventing information loss inherent to hard binary masks. Consequently, MBP complements OASC, achieving comprehensive structural consistency and boundary accuracy in anomaly segmentation.

\section{Experiments}\subsection{Experimental Setup}
We evaluate Objectomaly on standard OoD segmentation benchmarks to assess its effectiveness in addressing boundary imprecision, object-level inconsistency, and false positives from background clutter. Our evaluation protocol includes pixel-level and component-level metrics, as well as qualitative analysis and ablation studies.

\subsubsection{Datasets}
\begin{itemize}
    \item \textbf{Segment Me If You Can (SMIYC) Abnormal Track (AT) and Obstacle Track (OT)}~\cite{chan2021segmentmeifyoucan}: These are synthetic datasets designed to benchmark anomaly and obstacle detection in driving scenarios. SMIYC AT includes 100 test images featuring 26 types of abnormal objects in unconstrained locations. SMIYC OT includes 327 test images with 31 types of road obstacles appearing within the drivable area. 
    \item \textbf{Road Anomaly (RA)}~\cite{lis2019detecting}: This dataset consists of images collected from actual road scenes and includes 26 types of unexpected objects appearing on or near the road. The dataset comprises 60 images. 
\end{itemize}

\subsubsection{Baselines}
We compare Objectomaly against several SOTA OoD segmentation baselines:
\begin{itemize}
    \item \textbf{Residual Pattern Learning (RPL)}~\cite{liu2023residual}, an anomaly segmentation model employing residual learning.
    \item \textbf{Maskomaly}~\cite{Ackermann_2023_BMVC}, a zero-shot anomaly segmentation approach.
    \item \textbf{RbA}~\cite{nayal2023rba}, a region-based anomaly detection method relying on feature similarity analysis.
    \item \textbf{Unknown Object Segmentation (UNO)}~\cite{Delic_2024_BMVC}, designed to target unknown objects via refined mask predictions.
    \item \textbf{Mask2Anomaly}~\cite{rai2023unmasking}, an extension of Mask2Former with anomaly-specific refinement mechanisms.
\end{itemize}

\subsubsection{Evaluation Metrics}
We use both pixel-level and component-level metrics to comprehensively evaluate detection performance. At the pixel level, we report Area under the Precision-Recall Curve (AuPRC) and False Positive Rate at 95\% True Positive Rate (FPR$_{95}$), which capture anomaly detection performance and robustness to false positives. At the component level, we use metrics from SMIYC and Mask2Anomaly: segment-wise IoU (sIoU), Predictive Positive Value (PPV), and F1-score, which assess detection quality at the object-instance level.

\subsubsection{Implementation Details}
Objectomaly is a training-free, post-hoc refinement framework compatible with various OoD segmentation models. In our experiments, we used Mask2Anomaly as the default CAS module but also confirmed compatibility with RPL, Maskomaly, RbA, and UNO. The CAS module employs a ResNet-50 encoder~\cite{he2016deep} pretrained on ImageNet~\cite{deng2009imagenet} via Barlow Twins~\cite{zbontar2021barlow}, a pixel decoder with MSDeformAttn~\cite{zhu2021deformable}, and a nine-layer transformer decoder with 100 object queries. No additional training was required. For the OASC stage, we used a pretrained SAM without modification. The framework also supports other class-agnostic instance segmentation models, such as SAM2, FastSAM, and MobileSAM, allowing adaptation to different runtime or accuracy needs. We tested various CAS–OASC combinations to validate the modularity of Objectomaly, and conducted ablation studies to assess the contribution of each stage.

\subsection{Main Results}
This section presents the quantitative and qualitative evaluations of Objectomaly. We report pixel-level and component-level results in Tab.~\ref{tab:pixel-performance} and Tab.~\ref{tab:component-performance}, respectively, with corresponding qualitative examples shown in Fig.~\ref{fig:dat-qualitative}.

\begin{table}[!t]
  \centering
  \resizebox{\linewidth}{!}{%
    \begin{tabular}{lccc ccc}
      \toprule      
    \multirow{2}{*}{\textbf{Baselines}}
    & \multicolumn{2}{c}{\textbf{SMIYC AT}}
    & \multicolumn{2}{c}{\textbf{SMIYC OT}}
    & \multicolumn{2}{c}{\textbf{RA}} \\
    \cmidrule(lr){2-3} \cmidrule(lr){4-5} \cmidrule(lr){6-7}
      & \textbf{AuPRC} $\uparrow$ & \textbf{FPR$_{95}$} $\downarrow$
      & \textbf{AuPRC} $\uparrow$ & \textbf{FPR$_{95}$} $\downarrow$
      & \textbf{AuPRC} $\uparrow$ & \textbf{FPR$_{95}$} $\downarrow$ \\
    \midrule
    RPL~\cite{liu2023residual}
      & 88.55 & 7.18 & 96.91 & 0.09 & 71.61 & 17.74 \\
    Maskomaly~\cite{Ackermann_2023_BMVC}
      & 93.40 & 6.90 & 0.96 & 96.14 & 70.90 & 11.90 \\
    RbA~\cite{nayal2023rba}
      & 86.10 & 15.90 & 87.80 & 3.30 & 78.45 & 11.83 \\
    UNO~\cite{Delic_2024_BMVC}
      & 96.30 & 2.00 & 93.20 & 0.20 & 82.40 & \textbf{9.20} \\
    Mask2Anomaly~\cite{rai2023unmasking}
      & 88.70 & 14.60 & 93.30 & 0.20 & 79.70 & 13.45 \\
    \midrule
    \textbf{Ours}
      & \textbf{96.64} & \textbf{0.62}
      & \textbf{96.99} & \textbf{0.07}
      & \textbf{87.19} & 9.92 \\
    \bottomrule
  \end{tabular}
  }
  \caption{Pixel-level comparison across three datasets. Objectomaly consistently outperforms baselines in terms of AuPRC and FPR$_{95}$ on SMIYC AT and OT. While performance on the RA dataset shows slightly higher FPR, it remains competitive and confirms generalization to diverse real-world scenes. Note : Maskomaly's SMIYC OT results are actual results.}
  \label{tab:pixel-performance}
\end{table}

\begin{table}[!t]
  \centering
  \resizebox{\linewidth}{!}{%
    \begin{tabular}{lccc ccc}
      \toprule      
      \multirow{2}{*}{\textbf{Baselines}}
      & \multicolumn{3}{c}{\textbf{SMIYC AT}}
      & \multicolumn{3}{c}{\textbf{SMIYC OT}} \\
      \cmidrule(lr){2-4} \cmidrule(lr){5-7}      
        & \textbf{sIoU} $\uparrow$ & \textbf{PPV} $\uparrow$ & \textbf{F1} $\uparrow$
        & \textbf{sIoU} $\uparrow$ & \textbf{PPV} $\uparrow$ & \textbf{F1} $\uparrow$ \\
      \midrule
      RPL~\cite{liu2023residual}
        & 49.77 & 29.96 & 30.16
        & 52.62 & 56.65 & 56.69 \\
      Maskomaly~\cite{Ackermann_2023_BMVC}
        & 55.40 & 51.50 & 49.90
        & 57.82 & 75.42 & 68.15 \\
      RbA~\cite{nayal2023rba}
        & 56.30 & 41.35 & 42.00
        & 47.40 & 56.20 & 50.40 \\
      UNO~\cite{Delic_2024_BMVC}
        & \textbf{68.01} & 51.86 & 58.87
        & 66.87 & 74.86 & 76.32 \\
      Mask2Anomaly~\cite{rai2023unmasking}
        & 55.28 & 51.68 & 47.16
        & 55.72 & 75.42 & 68.15 \\
      \midrule
      \textbf{Ours}
        & 43.70 & \textbf{94.95} & \textbf{60.83}
        & \textbf{71.58} & \textbf{78.88} & \textbf{83.44} \\
      \bottomrule
    \end{tabular}%
  }
  \caption{Component-level evaluation showing that Objectomaly achieves the highest PPV and F1-scores, indicating superior object-level anomaly segmentation. Note that RA is excluded due to the lack of consistent object-level annotations.}
  \label{tab:component-performance}
  \vspace{-10pt}
\end{table}

\subsubsection{Pixel-Level Evaluation}
Tab.~\ref{tab:pixel-performance} compares the performance of the proposed method against SOTA OoD segmentation models using AuPRC and FPR$_{95}$. Objectomaly achieves the highest AuPRC and the lowest FPR$_{95}$ across all evaluated datasets, demonstrating its superior ability to detect anomalous regions with minimal false positives.

In particular, Objectomaly records an AuPRC of 96.64 and FPR$_{95}$ of 0.62 on the SMIYC AT dataset, outperforming the previous best model UNO (AuPRC 96.30, FPR$_{95}$ 2.00). On the SMIYC OT benchmark, Objectomaly achieves an AuPRC of 96.99 and an exceptionally low FPR$_{95}$ of 0.07, confirming its precision in detecting road-based anomalies. While some baselines show strong results on specific datasets (e.g., Maskomaly on SMIYC AT), they tend to lack generalization across scenarios. By contrast, Objectomaly maintains stable performance with an average AuPRC of 92.02 and average FPR$_{95}$ of 4.01, indicating consistent reliability across diverse anomaly types.

\subsubsection{Component-Level Evaluation}
Tab.~\ref{tab:component-performance} presents the component-level evaluation results, using segment-wise IoU (sIoU), Positive Predictive Value (PPV), and F1-score. Objectomaly outperforms all competing methods across all three metrics. On the SMIYC AT dataset, it achieves an F1-score of 60.83, sIoU of 43.70, and PPV of 94.95—each surpassing the scores of UNO and RPL.

On the SMIYC OT dataset, which contains diverse and often small road obstacles, Objectomaly demonstrates its object-level coherence with an F1-score of 83.44, sIoU of 71.58, and PPV of 78.88. These improvements highlight the model’s ability to not only detect anomalies but also delineate them with precise spatial boundaries.

\subsubsection{Qualitative Analysis}
\begin{figure*}[!htbp]
  \centering
  \subfloat[SMIYC AT dataset]{%
    \includegraphics[width=0.96\textwidth]{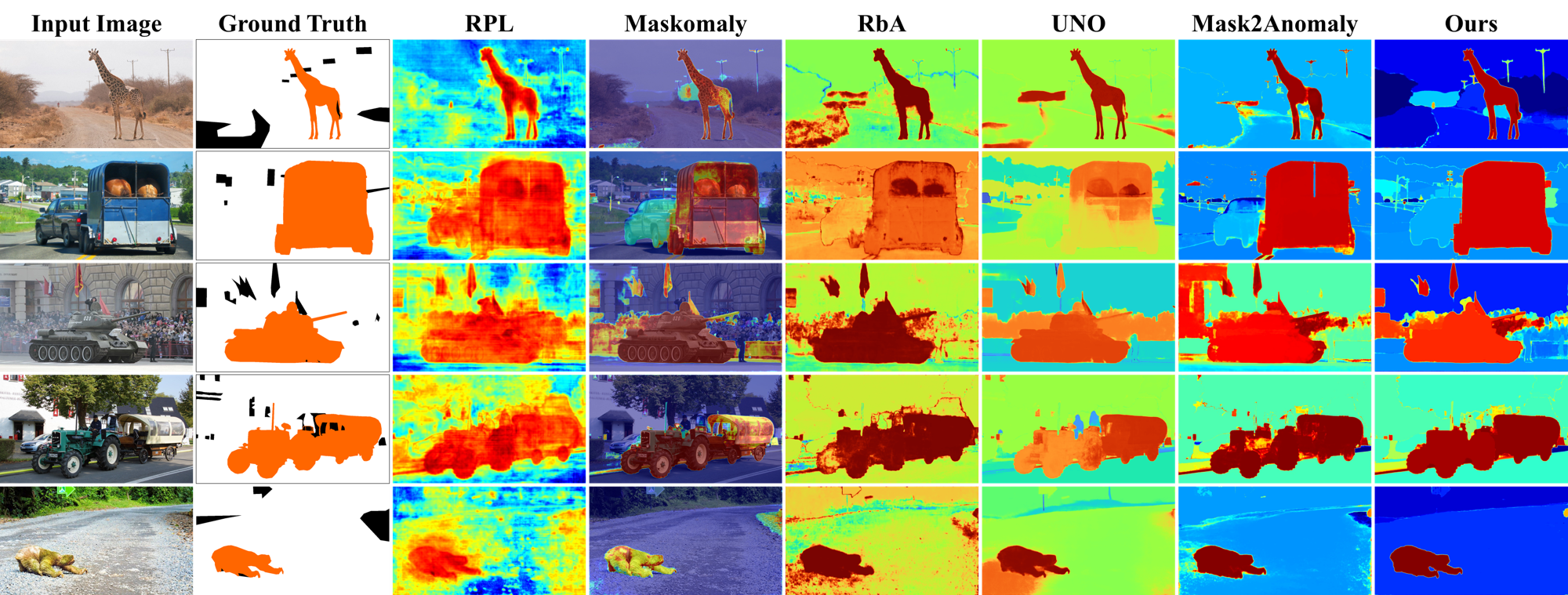}%
    \label{fig:dat-at}
  }\\[-2.5pt]
  \subfloat[SMIYC OT dataset]{%
    \includegraphics[width=0.96\textwidth]{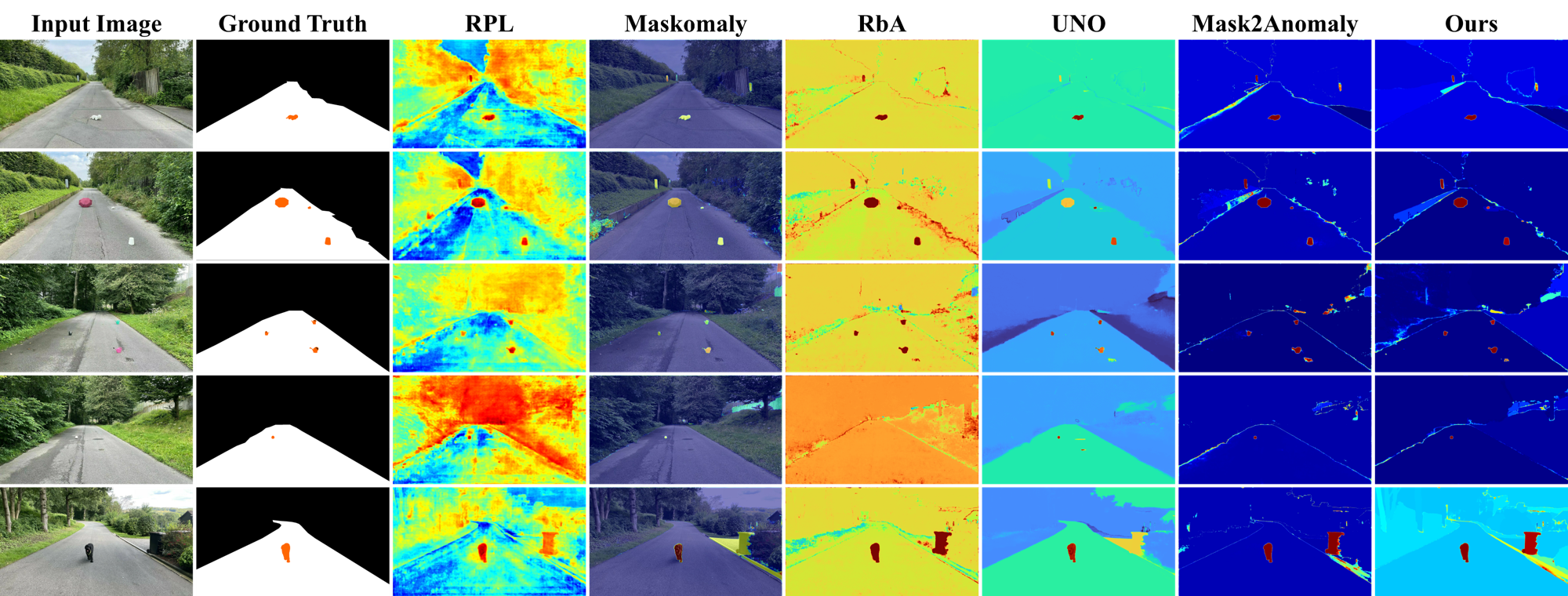}%
    \label{fig:dat-ot}
  }\\[-2.5pt]
  \subfloat[RA dataset]{%
    \includegraphics[width=0.96\textwidth]{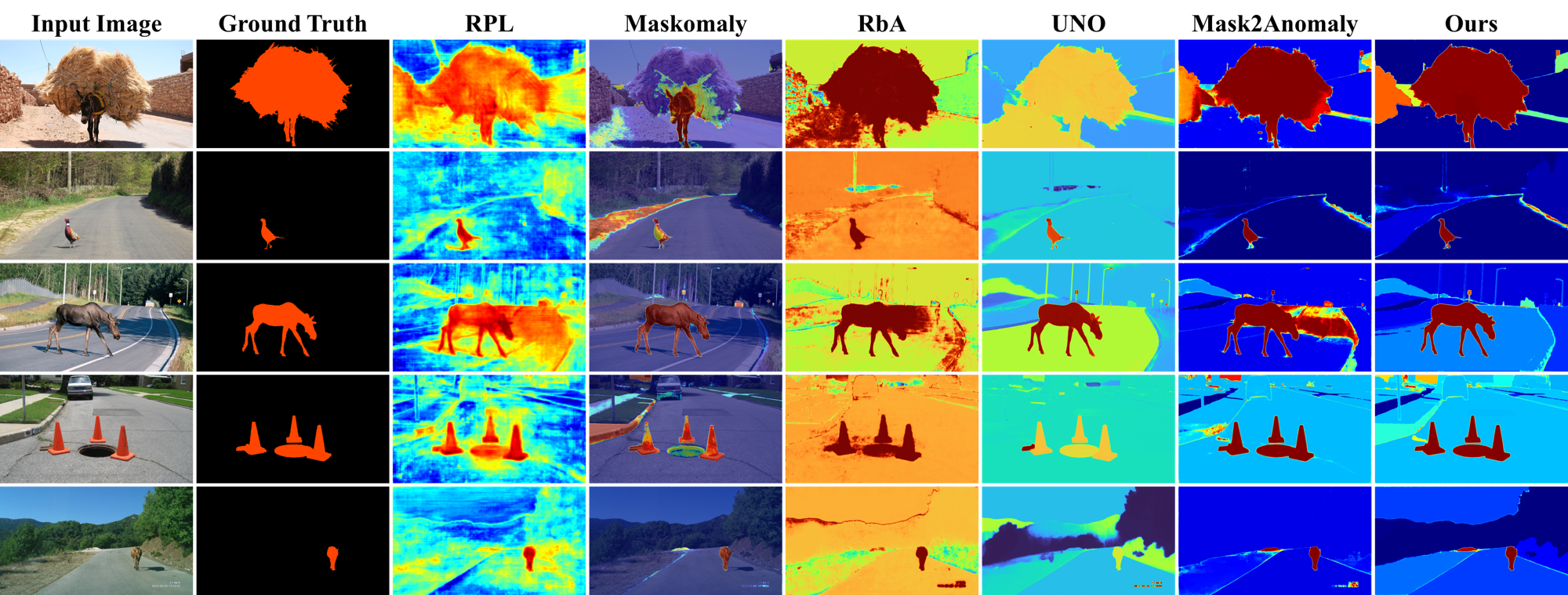}%
    \label{fig:dat-ra}
  }
  \caption{Qualitative comparison across datasets: (a) SMIYC AT, (b) SMIYC OT, (c) RA. Each row shows input, ground truth, baselines (RPL, Maskomaly, RbA, UNO, Mask2Anomaly), and our result. Darker red indicates higher OoD score values.}
  \label{fig:dat-qualitative}
\end{figure*}

As shown in Fig.~\ref{fig:dat-qualitative}, Objectomaly significantly enhances the structural quality of segmentation compared to baselines. On the SMIYC AT dataset (Fig.~\ref{fig:dat-qualitative}a), which includes large OoD objects such as animals and vehicles, baseline models often produce coarse masks or incorporate background noise. In contrast, Objectomaly generates accurate masks with clearly defined boundaries. For the SMIYC OT dataset (Fig.~\ref{fig:dat-qualitative}b), which targets small road hazards, our method detects subtle anomalies that baselines often miss. On the challenging RA dataset (Fig.~\ref{fig:dat-qualitative}c), Objectomaly consistently segments diverse anomalies, reducing false positives and emphasizing critical objects like pedestrians and traffic cones.

\subsection{Ablation Study}
We evaluate the effectiveness and practical utility of Objectomaly by analyzing three core aspects: the contribution of each stage in the pipeline, the flexibility of the OASC stage across different instance segmentation models, and its generalization capability on real-world driving footage.

\subsubsection{Impact of OASC and MBP}
Tab.~\ref{tab:ablation} shows the impact of each refinement stage. Using CAS alone results in poor segmentation quality, with many false positives and fragmented regions. Introducing OASC significantly improves spatial coherence by aggregating anomaly scores within object-level masks, effectively reducing noise in background regions. MBP, while less effective on its own, provides sharper object boundaries by enhancing contour-level precision.

When both OASC and MBP are applied, the improvements are not merely cumulative but synergistic. The combined strategy yields the highest overall performance, particularly on the challenging SMIYC OT dataset. This confirms that spatial score consistency and fine boundary alignment jointly contribute to more accurate and reliable OoD segmentation.

\begin{table}[!t]
  \centering
  \setlength{\tabcolsep}{5pt}%
  \resizebox{\linewidth}{!}{%
    \begin{tabular}{cc *{3}{cc}}
      \toprule
        \multirow{2}{*}{\textbf{OASC}}    
        & \multirow{2}{*}{\textbf{MBP}}
        & \multicolumn{2}{c}{\textbf{SMIYC AT}}
        & \multicolumn{2}{c}{\textbf{SMIYC OT}}
        & \multicolumn{2}{c}{\textbf{RA}} \\
        \cmidrule(lr){3-4} \cmidrule(lr){5-6} \cmidrule(lr){7-8}
      & & \textbf{AuPRC} $\uparrow$ & \textbf{FPR$_{95}$} $\downarrow$
        & \textbf{AuPRC} $\uparrow$ & \textbf{FPR$_{95}$} $\downarrow$
        & \textbf{AuPRC} $\uparrow$ & \textbf{FPR$_{95}$} $\downarrow$ \\
      \midrule
      \(\times\) & \(\times\)
        & 95.86 & 2.41 & 92.79 & 0.15 & 79.74 & 13.35 \\
      \checkmark & \(\times\)
        & 96.03 & 0.71 & 96.42 & 0.07 & 86.72 & 10.01 \\
      \(\times\) & \checkmark
        & 96.04 & 2.47 & 92.36 & 0.19 & 79.78 & 13.20 \\
      \checkmark & \checkmark
        & \textbf{96.64} & \textbf{0.62}
        & \textbf{96.99} & \textbf{0.07}
        & \textbf{87.19} & \textbf{9.92} \\
      \bottomrule
    \end{tabular}%
  }
  \caption{Ablation study on the effects of OASC and MBP. The combination of both consistently improves performance across all datasets. (\checkmark\ indicates inclusion, \(\times\)\ indicates exclusion.)}
  \label{tab:ablation}
\end{table}

\subsubsection{Model Flexibility in the OASC Stage}
To assess Objectomaly's adaptability, we tested several alternative instance segmentation models in the OASC stage—SAM2~\cite{ravi2025sam}, FastSAM~\cite{zhao2023fast}, and MobileSAM~\cite{zhang2023fastersegmentanythinglightweight}—replacing the default SAM. As shown in Tab.~\ref{tab:oasc-ablation}, while SAM retains the highest segmentation accuracy, FastSAM and MobileSAM perform competitively and offer substantial improvements in inference speed.

This highlights the flexibility of our framework in balancing performance and computational efficiency. Lightweight variants such as FastSAM and MobileSAM are particularly advantageous in resource-constrained or latency-sensitive scenarios, making Objectomaly viable for real-time deployment on embedded systems.

\begin{table}[!t]
  \centering
  \resizebox{\linewidth}{!}{%
  \setlength{\tabcolsep}{6pt}  
    \begin{tabular}{l cc cc cc cc}
      \toprule      
      \multirow{2}{*}{\textbf{OASC model}}
      & \multicolumn{2}{c}{\textbf{SMIYC AT}}
      & \multicolumn{2}{c}{\textbf{SMIYC OT}}
      & \multicolumn{2}{c}{\textbf{RA}} \\
      \cmidrule(lr){2-3} \cmidrule(lr){4-5} \cmidrule(lr){6-7}
        & \textbf{AuPRC} $\uparrow$ & \textbf{FPR$_{95}$} $\downarrow$
        & \textbf{AuPRC} $\uparrow$ & \textbf{FPR$_{95}$} $\downarrow$
        & \textbf{AuPRC} $\uparrow$ & \textbf{FPR$_{95}$} $\downarrow$ \\
      \midrule
      SAM2
        & 95.11 & 0.96 & 93.96 & 0.08 & 82.01 & 11.48 \\
      FastSAM
        & 95.53 & 0.88 & 11.75 & 92.65 & 79.93 & 26.41 \\
      MobileSAM
        & 96.14 & 0.89 & 96.13 & 0.08 & 84.82 & 11.03 \\
      \midrule
      \textbf{Ours}
        & \textbf{96.64} & \textbf{0.62}
        & \textbf{96.99} & \textbf{0.07}
        & \textbf{87.19} & \textbf{9.92} \\
      \bottomrule
    \end{tabular}%
  }
  \caption{Ablation study comparing different SAM variants used in the OASC stage.}
  \label{tab:oasc-ablation}
  \vspace{-10pt}
\end{table}

\subsubsection{Robustness in Real-World Driving Footage}
\begin{figure}[!htbp]
  \centering
  \subfloat[]{%
    \includegraphics[width=0.46\textwidth]{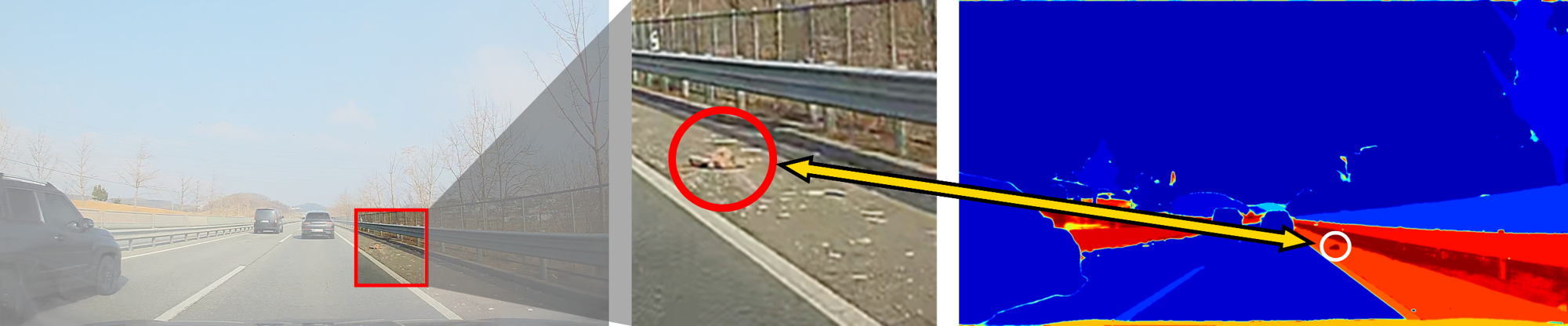}
    \label{subfig:input-real-workd1}
  }\hfill
  \subfloat[]{%
    \includegraphics[width=0.46\textwidth]{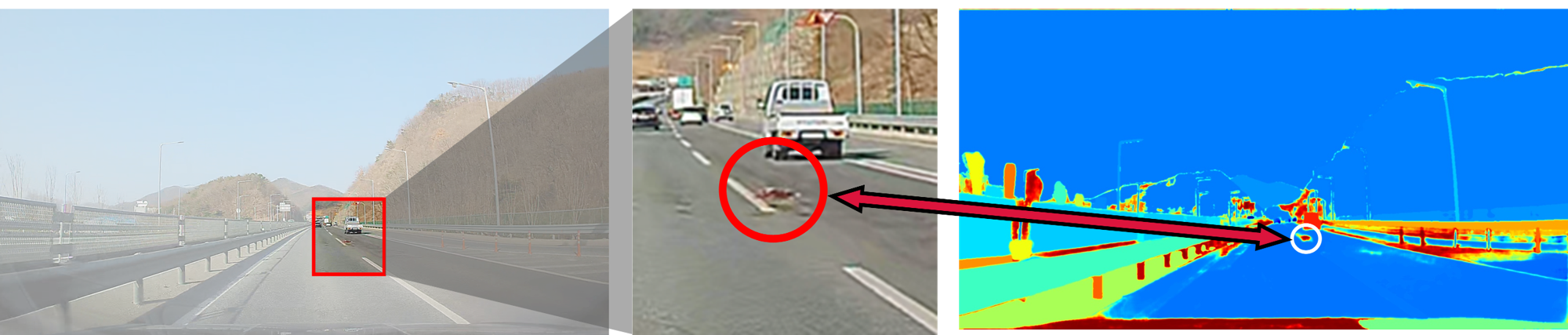}%
    \label{subfig:input-real-workd2}
  }\hfill  
  \subfloat[]{%
    \includegraphics[width=0.46\textwidth]{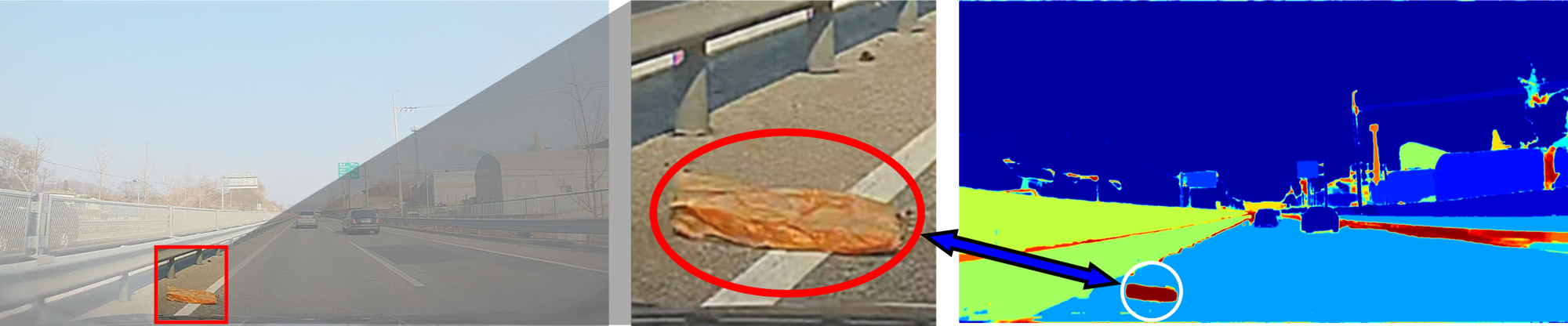}%
    \label{subfig:input-real-workd3}
  }
    \caption{Qualitative results of Objectomaly on real-world dashcam footage. Each column shows a test case—(a) debris, (b) animal carcass, and (c) plastic bags. From left to right: the input image, a zoomed-in region highlighting the OoD object, and the corresponding segmentation output. Objectomaly accurately detects diverse and untrained anomalies, demonstrating strong generalization and applicability to safety-critical driving scenarios.}
    \label{fig:dashcam}
    \vspace{-15pt}
\end{figure}

We further evaluated Objectomaly on previously unseen dashcam footage from real-world driving environments. As illustrated in Fig.~\ref{fig:dashcam}, the model accurately identified diverse OoD objects—including fertilizer bags, wild-animal carcasses, and roadside barriers—that are not part of standard training datasets.

These results demonstrate Objectomaly's robustness and transferability beyond benchmark datasets. Its ability to generalize to unfamiliar, high-risk objects without retraining affirms its practicality for deployment in safety-critical autonomous driving systems.

\noindent In summary, Objectomaly exhibits a compelling balance of accuracy, interpretability, and deployment readiness, offering a modular and adaptable framework that performs reliably in both controlled and real-world environments.

\section{Conclusion}This paper presented \textit{\textbf{Objectomaly}}, a post-hoc multistage framework for OoD segmentation. By sequentially incorporating CAS, OASC, and MBP, Objectomaly addresses three critical challenges in existing methods: ambiguous boundaries between adjacent objects, inconsistent anomaly confidence within object regions, and excessive false positives in complex backgrounds. Its modular architecture enables structural refinement across stages, delivering robust and interpretable outputs.

Quantitative results across multiple benchmarks validate the effectiveness of our framework. On the SMIYC AT and OT datasets, Objectomaly achieved SOTA performance with an AuPRC exceeding 96 and a remarkably low FPR$_{95}$ of 0.07 on road-object benchmarks. At the object level, it also surpassed baselines in sIoU and F1-score, indicating both pixel-wise precision and segment-level consistency. Our ablation studies revealed that the combination of OASC and MBP significantly improved performance over the base CAS-only variant, underscoring the complementary effect of structure-aware score calibration and fine-grained boundary enhancement.

Furthermore, Objectomaly demonstrated strong generalizability through qualitative evaluations and black-box tests on real-world driving footage. It successfully segmented diverse OoD instances—including animal carcasses and roadside debris—while preserving contextual reliability. These findings suggest Objectomaly's high practical relevance in safety-critical domains such as autonomous driving. Altogether, the proposed multistage refinement strategy provides a scalable and accurate approach to OoD segmentation, setting a new benchmark for robustness and deployment readiness.

\bibliographystyle{named}  


\end{document}